\documentclass[conference]{IEEEtran}
\IEEEoverridecommandlockouts
\usepackage{cite}
\usepackage{amsmath,amssymb,amsfonts}
\usepackage{algorithmic}
\usepackage{graphicx}
\usepackage{textcomp}
\usepackage{xcolor}
\usepackage{comment}
\usepackage{soul}
\usepackage{glossaries}
\usepackage{tikz}
\usepackage{subcaption}
\usepackage{gensymb}
\def\BibTeX{{\rm B\kern-.05em{\sc i\kern-.025em b}\kern-.08em
    T\kern-.1667em\lower.7ex\hbox{E}\kern-.125emX}}
\newcommand{\revisioni}[1]{#1}
\begin{document}

\title{Towards an extension of Fault Trees in the Predictive Maintenance Scenario}

\author{\IEEEauthorblockN{Roberta De Fazio, Stefano Marrone, Laura Verde}
\IEEEauthorblockA{\textit{Dipartimento di Matematica e Fisica} \\
\textit{Università della Campania “Luigi Vanvitelli”}, Caserta, IT \\
\{roberta.defazio,stefano.marrone,laura.verde\}@unicampania.it}
\and
\IEEEauthorblockN{Vincenzo Reccia, Paolo Valletta}
\IEEEauthorblockA{\textit{Gematica S.r.l.} \\
\textit via Diocleziano 107 --- Naples, Italy \\
\{v.reccia,p.valletta\}@gematica.com}
}

\maketitle

\begin{abstract}
One of the most appreciated features of Fault Trees (FTs) is their simplicity, making them fit into industrial processes. As such processes evolve in time, considering new aspects of large modern systems, modelling techniques based on FTs have adapted to these needs. This paper proposes an extension of FTs to take into account the problem of Predictive Maintenance, one of the challenges of the modern dependability field of study. The paper sketches the Predictive Fault Tree language and proposes some use cases to support their modelling and analysis in concrete industrial settings. 
\end{abstract}


\newacronym{ai}{AI}{Artificial Intelligence}
\newacronym{arl}{ARL}{Association Rules Learning}
\newacronym{bn}{BN}{Bayesian Network}
\newacronym{bpmn}{BPMN}{Business Process Modeling Notation}
\newacronym{mde}{MDE}{Model Driven Engineering}
\newacronym{ci}{CI}{Critical Infrastructure}
\newacronym{cps}{CPS}{Cyber-Physical System}
\newacronym{dag}{DAG}{Directed Acyclic Graph}
\newacronym{dbn}{DBN}{Dynamic Bayesian Network}
\newacronym{dd}{DD}{Data-Driven}
\newacronym{dft}{DFT}{Dynamic Fault Tree}
\newacronym{dm}{DM}{Data Mining}
\newacronym{dt}{DT}{Digital Twin}
\newacronym{dts}{DTs}{Digital Twins}
\newacronym{fr}{FR}{Functional Requirement}
\newacronym{fta}{FTA}{Fault Tree Analysis}
\newacronym{ft}{FT}{Fault Tree}
\newacronym{iot}{IoT}{Internet of Things}
\newacronym{mb}{MB}{Model-Based}
\newacronym{ml}{ML}{Machine Learning}
\newacronym{nfi}{NFI}{Non-Functional Index}
\newacronym{nfr}{NFR}{Non-Functional Requirement}
\newacronym{pdft}{PdFT}{Predictive  Fault Tree}
\newacronym{pdm}{PdM}{Predictive Maintenance}
\newacronym{pm}{PM}{Process Mining}
\newacronym{pts}{PTs}{Petri Nets}
\newacronym{qn}{QN}{Queueing Network}
\newacronym{rft}{RFT}{Reparaible Fault Tree}
\newacronym{seft}{SEFT}{State-Event Fault Tree}
\newacronym{tsa}{TSA}{Time Series Analysis}
\newacronym{uml}{UML}{Unified Modeling Language}
\newacronym{darwinist}{DARWINIST}{aDversarial scenArios geneRation With dIgital twiNs In induSTry} 
\newacronym{as}{AS}{Association Rules}
\newacronym{lstm}{LSTM}{Long short term memory}
\newacronym{rul}{RUL}{Remaining Useful Life}
\newacronym{ocel}{OCEL 2.0}{Object Centric Event Log}
\newacronym{pn}{PN}{Petri Net}
\newacronym{bn}{BN}{Bayesian Network}
\newacronym{mtbf}{MTBF}{Mean Time Between Failures}
\newacronym{mttf}{MTTF}{Mean Time To Failures}
\newacronym{dsml}{DSML}{Domain Specific Modelling Language}

\section{Introduction}
\label{sec:intro}


In the context of Industry 4.0, the \gls{pdm} has gained ground, becoming one of the more interesting applications of \gls*{iot} and \gls*{ai}, from healthcare to railway ~\cite{01-14,01-16,10331740}. The need to ensure high reliability in critical systems is leading to more precise analysis techniques ~\cite{01-2}. Both \gls{mb} ~\cite{01-11-1, 01-5} and \gls{dd} ~\cite{01-10} approaches have been widely adopted to accomplish these issues.
Among \gls*{mb} methods, \glspl*{ft} are a solid formalism, widely accepted in both academia and industry.
One of the most appreciated results achieved by this formalism is its adaptability to the evolution of systems and analysis processes: different extensions of \gls{ft} can capture various aspects of the system to model, from repairability to failure sequences.
This research has the objective of improving the \gls*{ft} formalism by adapting it to the new challenge of \gls*{pdm}. The long-term goal is to join the benefits of \gls{dd} techniques (flexibility and adaptability to several contexts) with the transparency and explainability of \gls*{mb} methods. 
Other possible benefits to demonstrate are the necessity of smaller datasets since the presence of prior knowledge about the context ~\cite{01-9, DONG2017374} traditionally lacks in pure-\gls*{dd} approaches. 
Last, \gls*{mb} methods are most accepted in critical system validation and certification processes. This problem is emphasized in \glspl*{cps}: industrial settings would greatly enjoy the \gls{ml}-based applications, but the capability to inspect a model --- as in \gls{mb} contexts --- is of great importance.
The possibility of combining the points of strengths of both methods, ensuring better realism and higher interpretability of the results led to propose new hybrid techniques.
Few works in the literature, indeed, propose similar approaches.
In ~\cite{01-10}, the authors use a set of mathematical equations as \gls{mb} technique for evaluating the residuals and classifying the faults by using \glspl*{bn} while in ~\cite{01-13} after computing residuals, the authors use Support Vector Data Description classifier for ranking minimal diagnosis candidates. The approach proposed in ~\cite{01-17} provides a hybrid \gls{bn}, boosted with a symptom layer including nodes with residuals' features, nodes containing knowledge-based information and nodes composed of data-extracted features. Another approach, proposed in ~\cite{01-18}, exploits \gls{bn} formalism trained on faulty data fused with the reference model driven on normal conditions.
Some approaches are based on the usage of \glspl*{dbn} ~\cite{Ghahramani20019} which is a more powerful formalism than \glspl*{ft}, even if there is a trade-off between expressiveness and usability of formal languages ~\cite{10.1007_3-540-48249-0_27}. Some works presenting the usage of \glspl*{dbn} in \gls*{pdm} are: ~\cite{Foulliaron2014849}, where \glspl*{dbn} are used to learn system observable dynamics, combined with false error correction by the usage of Conditional Probability Table; ~\cite{Zeng2023}, which relies more on learning algorithms for \glspl*{dbn}, without having an explicit modelling phase; ~\cite{Hosamo2023}, where a \gls*{bn} model is obtained by the merging an Extreme Gradient Boosting approach with prior knowledge embedded into an ontology. One of the most promising bridging techniques between \gls*{dd} and \gls*{mb} is \gls{pm}. \gls*{pm} is a technique for extracting process models from the real data ~\cite{PMmanifesto}. To the best of our knowledge, only a few works propose a solution for fault prediction in critical infrastructure via \gls{pm} ~\cite{Gunther2022, Ming200966}.
This paper proposes the \gls*{pdft} language, based on \glspl*{seft} ~\cite{Kaiser20071521} and \glspl*{dft}~\cite{DurgaRao2009872}. The main goal is the improvement of the expressive power of the model itself to allow the design of more complex structures with several behaviours and the evaluation of the performances under these critical behaviours. The language is introduced by defining an abstract syntax, a tentative graphical notation, and some initial considerations about the semantics. This notwithstanding, the proposed formalism is supported by a vision aiming at merging \gls*{mb} and \gls{dd} approaches; in particular, the usage of three different \gls*{dd} techniques as \gls*{pm}, \gls*{tsa} and \gls*{arl}.
Concerning this brief review of the state of the art, the original contributions present in this research are: (1) introduction of the \gls*{pdft} formalism;  (2) definition of a usage approach combining \gls*{mb} and \gls*{dd} approaches.
The rest of the paper is structured as follows:  Section \ref{sec:formalism} introduces the \gls*{pdft} formalism while Section \ref{sec:usecases} reports a usage scenario of the language, described also by the support of a small example. Section \ref{sec:conclusions} ends the paper, reporting the progress of the research and addressing the next steps.
\section{The \gls*{pdft} Formalism}
\label{sec:formalism}

This section introduces the \gls*{pdft} formalism with the description of its abstract syntax, a graphical notation proposal and some hints on the semantics of the language. At the top level, a \gls*{pdft} model is characterised by a quadruple $<\mathcal{C}, \mathcal{E}, \mathcal{D}, \epsilon, \Phi>$.

Let $\mathcal{C} = \{c_1, c_2, \ldots, c_n\}$ be a set of \textbf{components}, each of which can be related to one or more \textbf{ports}, which could be input ports (i.e., $P_{I}^{c}$), involved in the representation of how a component reacts to external changes, and output ports (i.e., $P_{O}^{c}$), which describe how a component propagates inner changes to the outside. Ports are considered in this paper as boolean variables.

The ports are connected by a set of \textbf{events}. An event $e \in \mathcal{E}$ is an element of a relation between input and output ports: $\mathcal{E} \subseteq  P_{I} \times P_{O}$\footnote{$Let P_{I}$ be the union of all the $P_{I}^{c}$s and $P_{O}$ the union of all the $P_{O}^{c}$s.}. The function $\epsilon$ is responsible for assigning a \textbf{weight} to each event; $\epsilon: \mathcal{E} \longrightarrow [0,1] \subset \mathbb{R}$.

The novelty of the \gls*{pdft} formalism is constituted by \textbf{dynamics}, collected in the $\mathcal{D}$ set. The dynamics model physical processes --- described as signals --- that influence the behaviour of the components. Dynamic conditions, i.e., boolean predicates over the values of the dynamics, model how components react differently to the dynamics. As an example, let us consider two distinct physical components of a system, placed in the same room. They can react in terms of failure probability to such a temperature according to different laws; one may start failing when the temperature is over 37 C\degree, while the other may start showing an intermittent failure pattern over 50 C\degree. More formally, a dynamic is a real function over time $d \in \mathcal{D}\:|\:d: \mathbb{R} \longrightarrow \mathbb{R}$.

Describing the inner features of $\mathcal{C}$, each component $c \in \mathcal{C}$ is characterised by:
\begin{itemize}
    \item a set of $M_c$ \textbf{states}, $S^{c} = \{s^c_1,\ldots\ s^c_{M_c}\} \cup \{\bullet\} \neq \emptyset$. 
    \revisioni{Every component has at least the initial state, represented by the bullet};
    \item a \textbf{state priority} function assigning a priority to each state, $\phi^c:S^{c}\rightarrow\mathbb{N}$, which is represented by a natural number;
    \item a set of oriented $T^c$ \textbf{transitions}, as a relation over the Cartesian product of states $T^{c} \subseteq S^{c} \times S^{c}$. 
\end{itemize}

Before characterising each component, let us consider:
\begin{itemize}
    \item $\mathbb{P}(P_{I}^{c} \cup \mathcal{D})$, the set of the boolean predicates over both the dynamic values and input ports;
    \item $\mathcal{P}(P_O^c)$, the powerset of the output ports of the $c$ component.
\end{itemize}

Each component's transition is also characterised by the following functions:
\begin{itemize}
    \item \textbf{trigger}: let $\tau^c: T^c  \longrightarrow \mathbb{P}(P_{I}^{c} \cup \mathcal{D})$ associate to each transition a boolean predicate over ports as well as dynamics;
    \item \textbf{action}: let $\alpha^c: T^c \longrightarrow \mathcal{P}(P_O^c)$, representing which output ports of the components are set to \textit{true};
    \item \textbf{time}: let $\delta^c: T^c  \longrightarrow \mathbb{R}$ represent the time needed to accomplish the transition;
    \item \textbf{transition priority}, $\pi^c: T^c  \longrightarrow \mathbb{N}$;
    \item \textbf{probability}, $\rho^c: T^c  \longrightarrow [0,1] \subset \mathbb{R}$, supposed constant, it could change with time and other conditions.
\end{itemize}

In the end, let us define the \textbf{global alert} function as the $\Phi$ able to return the global priority/alert level of the model, according to the states in which the different components are. More formally, $\Phi: S^{c_1} \times S^{c_2} \times \ldots \times S^{c_n} \longrightarrow \mathbb{N}$ so that $\Phi(s_1,s_2,s_n) = \min\limits_{i \in \{1,\ldots,n\}} \phi^{c_i}(s_i)$.

To introduce how a \gls*{pdft} model can be used to understand the availability of a system, let's imagine that the initial condition of the model consists: all the dynamics start at $t = 0$, all the components start with their initial pseudo-states $\bullet$, and that all the ports are set to \textit{false}. 

With a step of $\Delta_T$ a continuous simulation starts and at each cycle:
\begin{enumerate}
    \item dynamics are updated, according to their implementations;
    \item for each component, the triggers of the outbound transitions regarding the current states, are evaluated;
    \item when more than a transition may fire, the subset with a higher transition priority is chosen and, inside this set, the transition to fire is chosen probabilistically, according to the probability function;
    \item the firing of a transition activates the output ports reported in the action function;
    \item events propagate, according to the probability weight, the truth of an output port to one or more input ports;
    \item at each cycle, the $\Phi$ function is evaluated to understand at which global alert level the model is set.
\end{enumerate}

\subsection{An example of \gls*{pdft} model}

\begin{figure}[h!]
\centering
\includegraphics[width=0.6\columnwidth]{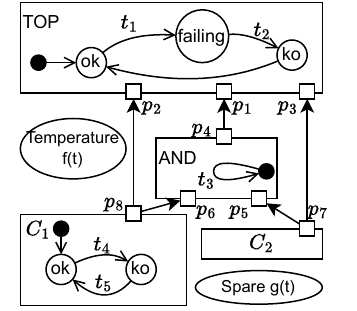}
\caption{A sample \gls{pdft} model and its notation} \label{fig:pdft.notation}
\end{figure}

Starting from this abstract formalisation, a familiar concrete notation with the objective of not diverging from the classical \gls*{ft}-related formalisms. Fig. \ref{fig:pdft.notation} represents a sample model expressed in \gls*{pdft}. 

It is worth highlighting some model aspects, showing the potentiality of the notation:
\begin{itemize}
    \item the \texttt{TOP} component contains three states whose priorities are growing: the condition $\varphi^{TOP}(ok) < \varphi^{TOP}(failing) <  \varphi^{TOP}(ko)$ means that $\Phi$ would raise an alarm at a certain level when the \texttt{TOP} component is in the state \texttt{failing} while it would raise a higher alarm in the \texttt{ko} state.
    \item while $t_2$ fires on the activation of the port $p_1$ (i.e., on the output of the \texttt{AND} gate), \revisioni{this behaviour is modelled by the trigger function $\tau^{TOP}(t_1) = p_2 \vee p_3$}

    \item the \texttt{AND} component implements a simple \textit{and} of the failures of the components $C_1$ and $C_2$. A possible representation is: $\tau^{AND}(t_3) = p_6 \wedge p_5$, $\alpha^{AND}(t_3) = \{p_4\}$   
    \item this model introduces two dynamics: $f(t)$, which models the temperature of the environment containing the system, and $g(t)$, which represents the stochastic process by which a spare part of the component $C_1$ is present in case of component failure. It is important to underline that the model is independent of how dynamics are specified: deterministic/stochastic closed analytic expression, real-world time series, and simulated data, are all acceptable for the formalism;
    \item $t_4$ can capture the dependency from the temperature dynamic, as an example, depicting that the $C_1$ fails if the temperature surpasses 37 C\degree, i.e., $\tau^{C_1}(t_4) = (f > 37)$. On the other hand, $t_5$ may capture a repair action from \texttt{ko} to \texttt{ok}, considering the possibility of repairing the block if there are some spare elements, $\tau^{C_1}(t_5) = (g \geq 1)$.
\end{itemize}

\section{\glspl*{pdft} Use Cases}
\label{sec:usecases}

The scope of this section is to frame the proposed formalism in a general use case, allowing not only an analysis process utilizing simulation but also enhancing the modelling process by integrating \gls*{mb} and \gls*{dd} techniques.

One of the concrete applications of the proposed language is railway network monitoring and management. Railways are traditionally considered a \gls{ci} since they provide essential social services. Modern railway systems are characterized by the massive adoption of \glspl*{cps} and sensing devices based on the \gls{iot} paradigm. \glspl*{ci} are composed of different and heterogeneous (sub-)systems: physical where sensors and actuators play a central role, and digital which are usually managed by a monitoring centre and by a communication network. Due to their large and heterogeneous nature, these systems are often affected by unexpected failures, making fault identification and diagnosis processes challenging to run: the presence of a formalism as \gls*{pdft} allows more flexible modelling and analysis approaches. Furthermore, \gls*{pdm} is something to pursue to improve the availability, keeping the overall ownership cost acceptable.

\begin{figure}[h!]
\centering
\includegraphics[scale=0.3]{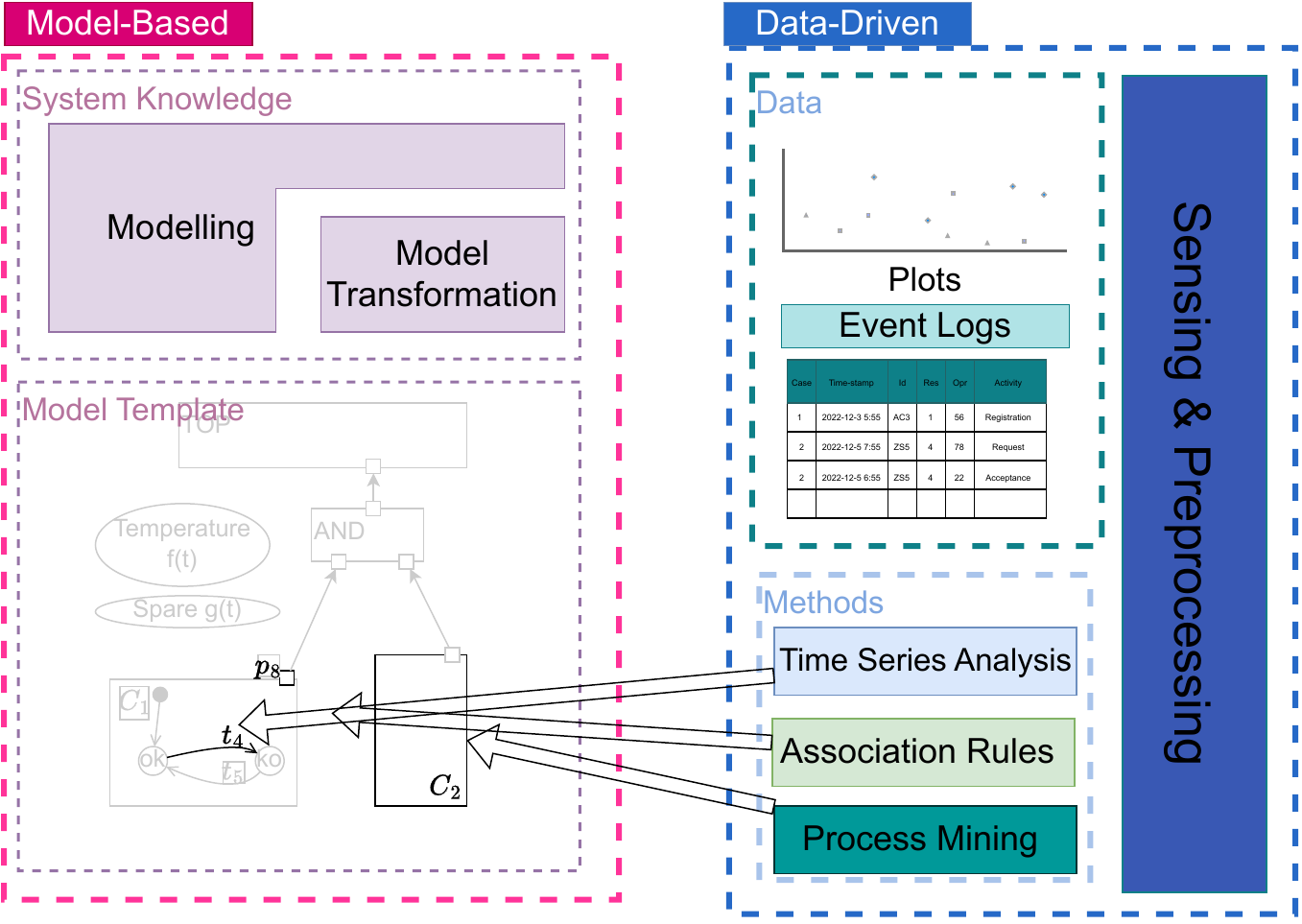}
\caption{An integrated \gls*{mb}-\gls*{dd} approach} \label{fig:usecase}
\end{figure}

Fig. \ref{fig:usecase} proposes a \gls*{mb}-\gls*{dd} integrated approach.

Through the usage of system knowledge, a modeller creates a model of the system. Two general strategies could be followed: create a model in \gls*{pdft} formalism, rely on a mode-transformation and on \gls*{mde} principles. The objective is not in creating a full model, but instead in creating a \textbf{model template}, i.e. a model where consolidated parts (i.e., the frozen spots, in grey in the figure \ref{fig:usecase}) are designed with changeable parts (i.e., the hot spots, in black in the fig. \ref{fig:usecase}). In the figure \ref{fig:usecase} is reported a simplified version of the example in Fig. \ref{fig:pdft.notation} as an example of \textbf{model template}. The second part of the approach is supported by \gls*{dd} methods. After a \textbf{Sensing \& Preprocessing} stage, data is available both in the form of plots and event logs. The first set of data is related to continuous variables (e.g., temperature, humidity, etc.), while the second ones are related to discrete-time events (e.g., faults, repair events). On such data, different \textbf{methods} could be adopted to improve the model template by “filling” its hot spots. Some of these methods and their possible applications are here described.

\paragraph*{Time Series} times series analysis techniques could be used to infer useful information. Two possible applications are: (1) simple interleaving statistical analysis to retrieve, for example, repair interval times (e.g., $\delta^{C_1}(t_5)$), (2) multidimensional techniques could be used to infer conditions of failures (i.e., by extracting conditions on dynamics values and their probabilities, computing proper trigger and probability functions for some transitions).

\paragraph*{Association Rules} this unsupervised technique is traditionally used to infer novel knowledge and to discover correlations in a dataset. The application of this technique to our case can be to understand that a failure of a component can cause the failure or a change of a state into another component. As an example, if a new is found by a \gls*{arl} algorithm saying that $(C_1 == ko \wedge TOP = ok) \Rightarrow TOP == ko$ with confidence 0.4, then: a new input port $p_x$ of \texttt{TOP} a new event \textit{e} from $p_8$ to $p_x$ with $\epsilon(e) = 0.4$ can be created and a new transition from \textit{ok} to \textit{ko} of \texttt{TOP} can be created with $p_x$ as a trigger. 

\paragraph*{Process Mining} the most complex scenario here presented can exploit the power of this \gls*{ml} technique to infer from the event log the inner failure process of a component. In this way, component behaviours could be defined from scratch or refined by discovering further transitions or states.

Table \ref{tab:mapping} reports a preliminary mapping of the possible \gls*{dd} techniques to the different aspects of the \gls*{pdft} formalism.

\begin{table}[!ht]
    \centering
    \caption{Mapping between \gls*{pdft} elements and \gls*{dd} methods}
    \label{tab:mapping}
    \begin{tabular}{|c|c|c|c|c|}
    \hline
        \textbf{Syntax Element} & \textbf{\gls{tsa}} & \textbf{\gls{arl}} & \textbf{\gls{pm}} \\ \hline
        ports & ~ & X & X \\ \hline
        weights & ~ & X & ~ \\ \hline
        triggers & X & X & ~ \\ \hline
        states & ~ & ~ & X \\ \hline
        transitions & ~ & X & X \\ \hline
        actions & ~ & X & ~ \\ \hline
        times & X & ~ & ~ \\ \hline
    \end{tabular}
\end{table}

\section{Perspectives and Future Work}
\label{sec:conclusions}


This paper has presented an extension of the well-known \glspl*{ft}: the \glspl*{pdft}. The objective of introducing another formalism is twofold: (1) the formalism can handle different aspects of a system failure process, from traditional logical gates to sequences of events and to repairable mechanisms; dynamics are a way to model the dependency of failures and anomalous states from environmental data; (2) the formalism can adapt to different \gls{dd} mechanisms, able to improve the adherence of a \gls*{pdft} model to reality, providing support to the modeller. 
\revisioni{Currently, the language has been defined from an abstract point of view and requires the definition of a concrete notation (textual and/or graphical). The efforts will be focused on a detailed description of the semantics of \gls*{pdft}.
In parallel, we are working on an implementation of the model leveraging \gls{dsml} expressive power. Among the platforms providing \gls{dsml} tooling construction, MPS\footnote{https://www.jetbrains.com/mps/} is a projectional editor not requiring special skills in grammar-based parser and and graphical model editors.
The aim is to explain what a \gls*{pdft} model means and how it can be analysed by translating each element of the language into another well-known language.
To pursue this goal, we detect Generalised Stochastic Petri Nets as a target language for the analysis. Previous transformation from \glspl*{ft}, \glspl*{rft} and \glspl*{dft} are already present in the literature, e.g., \cite{Raiteri2004659}. Lastly, Python-based simulator, which enables “testing” \gls*{pdft}-based models, is under finalization. This will allow us to build a case study based on a real scenario, on which define the \textbf{model template} and complete all the previous points to let the application of the proposed \gls*{dd} methods.}


\section*{Acknowledgment}

\footnotesize
The research has received funding by the  \gls*{darwinist} project, funded by Universit\'a della Campania “Luigi Vanvitelli”, D.R. 834 del 30/09/2022. The work of Laura Verde is granted by the  “Predictive Maintenance Multidominio (Multidomain predictive maintenance)” project, PON "Ricerca e Innovazione" 2014-2020, Asse IV "Istruzione e ricerca per il recupero"-Azione IV.4-"Dottorati e contratti di ricerca su tematiche dell'innovazione" programme CUP: B61B21005470007.  The work of Roberta De Fazio is granted by PON Ricerca e Innovazione 2014/2020 MUR — Ministero dell’Universit\`a e della Ricerca (Italy) — with the PhD program XXXVII cycle D.M. N.1061 "Dottorati e contratti di ricerca su tematiche dell'Innovazione".

\end{document}